\documentclass[sigconf]{acmart}

\usepackage{amsmath,amsfonts}
\usepackage{algorithmic}
\usepackage{graphicx}
\usepackage{textcomp}
\usepackage{hhline}
\usepackage{booktabs}
\usepackage{xcolor}
\usepackage{array}
\usepackage[T1]{fontenc}
\usepackage[utf8]{inputenc}
\usepackage{makecell}
\usepackage{multirow}
\setcopyright{acmlicensed}

\begin{document}

\title{HORAE: an annotated dataset of books of hours}

\author{M\'elodie Boillet}
\affiliation{%
 \institution{TEKLIA Paris}
 \institution{LITIS, Universit\'e de Rouen-Normandie}
 \country{France}
 }

\author{Marie-Laurence Bonhomme}
\affiliation{%
  \institution{TEKLIA Paris}
  \country{France}
 }

\author{Dominique Stutzmann}
\affiliation{%
  \institution{IRHT-CNRS, Paris}
  \country{France}
 }

\author{Christopher Kermorvant}
\affiliation{%
  \institution{TEKLIA Paris}
  \institution{LITIS, Universit\'e de Rouen-Normandie}
  \country{France}
}

\copyrightyear{2019}
\acmYear{2019}
\acmConference[HIP '19]{The 5th International Workshop on Historical Document Imaging and Processing}{September 20--21, 2019}{Sydney, NSW, Australia}
\acmPrice{15.00}
\acmDOI{10.1145/3352631.3352633}
\acmISBN{978-1-4503-7668-6/19/09}

\begin{abstract}
We introduce in this paper a new dataset of annotated pages from books of hours, a type of handwritten prayer books owned and used by rich lay people in the late middle ages. The dataset was created for conducting historical research on the evolution of the religious mindset in Europe at this period since the book of hours represent one of the major sources of information thanks both to their rich illustrations and the different types of religious sources they contain. We first describe how the corpus was collected and manually annotated then present the evaluation of a state-of-the-art system for text line detection and for zone detection and typing. The corpus is freely available for research.

\end{abstract}



\maketitle

\renewcommand{\shortauthors}{Boillet and Bonhomme, et al.}

\section{Introduction}

Automatic layout analysis of historical documents is very challenging due to the tremendous variety of documents produced over the last centuries. If we restrict the scope to illuminated medieval manuscripts, the number of concerned volumes is still very important but one can expect more similarities between them, at least regarding the types of elements found on the pages and their layout. Recognizing the text in medieval manuscripts has been a challenge for the last decades, but most of the systems focus on text line recognition and often in a single manuscript\cite{Fischer2011,Fischer2012}. System for automatic layout analysis are usually tested on a very limited number of manuscripts, for example, a single one for \cite{Grana2009} and six for \cite{Yang2017}.

Recently, international evaluations have been organized for medieval manuscripts layout analysis\cite{Mehri_2017}, providing a large number of annotated pages but from a restricted number of manuscripts (4,436 images from 11 books for \cite{Mehri_2017}).
As part of the HORAE research project \cite{dh_stutzmann2019}, aiming at studying the text of medieval devotional manuscripts, a large-scale annotated dataset was needed.
In this paper, we describe the HORAE corpus,  a corpus of annotated pages from books of hours collected for the study of their structure both from a layout and textual point of view. We first describe how the book of hours was selected in libraries from different countries and the composition of the corpus. We then describe how a sample of pages to be annotated has been selected and how it was annotated. Finally, we evaluate a state-of-the-art system for automatic layout analysis on the annotated pages. The dataset is freely accessible at \url{https://github.com/oriflamms/HORAE/}.

\begin{figure}[t!]
 \centering
 \includegraphics[width=0.5\textwidth]{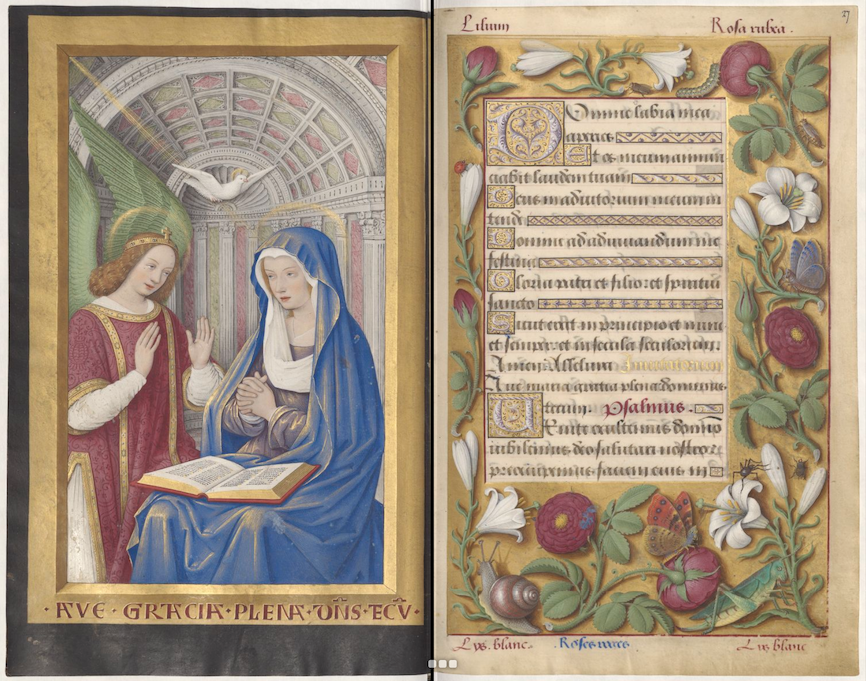}
 \caption{Two pages from the book of hours in latin "Grandes Heures d'Anne de Bretagne", Bibliothèque nationale de France. Département des Manuscrits, Latin 9474.}
\end{figure}

\section{Description of the corpus}

\subsection{Books of hours}

 In the late Middle Ages, books of hours were owned and used by the laity as personal prayerbooks. They enjoyed a wide diffusion and more than ten thousand of them survive today in libraries, museums, and in private hands. They have been described as the \textit{medieval best-seller}, but they show a great diversity both in the text and in the illustrations. Books of hours are a major historical source for the historical study of arts and painting since most of them are decorated with pictures (miniatures), decorated borders and initials. They are also crucial to understand the devotion and religious mindsets in medieval Europe because their numerous and very diverse texts are witnesses to customization practices and personal involvement in reading. The textual content, barely studied by scholars until now, is the object of the interdisciplinary research project HORAE (Hours: Recognition, Analysis, Edition)\cite{horae}, encompassing HTR, Natural Language Processing and Humanities. To process automatically at a large scale such a heterogeneous collection, we recognized the need to create an annotated dataset of pages from multiple books of hours.

\subsection{Sources}
The corpus encompasses 500 fully digitized manuscripts\footnote{List as of 12 Oct. 2018: file 500\_MSS.csv available at  \url{https://github.com/oriflamms/HORAE/}.}. They have been manually selected as part of the HORAE project. The present list is based on a larger census of books of hours\cite{Stutzmann19}
and restricted to digitized manuscripts fully available through the IIIF International Image Interoperability Framework\footnote{\url{https://iiif.io/}}. The number of manuscripts and images for each provider are given in Table \ref{providers}. The main providers are the digital libraries BVMM (Bibliothèque virtuelle des manuscrits médiévaux) and Gallica, maintained respectively by Institut de Recherche et d'Histoire des Textes (IRHT-CNRS) and the French National Library.

\begin{table}[ht]
\caption{Number of Manuscripts and images in the dataset for each provider.}
\label{providers}
\centering
\begin{tabular}{lllrr}
    \toprule
\textbf{\sc{Provider}} & \textbf{\sc{City}} & \textbf{\sc{MSS}} & \textbf{\sc{Images}}\\
    \midrule
    ugent.be & Gent & 1 & 142 \\
       \midrule
\textbf{BVMM}& &  \textbf{275} & \textbf{41902}\\
\url{https://bvmm.irht.cnrs.fr}& $\leq8$ MSS & 114 \\
& Angers & 21 \\
& Autun	& 12 \\
& Auxerre & 10 \\
& Beaune & 15 \\
& Chantilly & 30 \\
& Nantes & 18 \\
& Paris & 17 \\
& Rennes & 23 \\
& Toulouse & 15 \\
\midrule
\textbf{Gallica} & \textbf{Paris, BNF} & \textbf{183} & \textbf{52923} \\
 https://gallica.bnf.fr& -- Arsenal & 38 \\
 & -- Manuscrits &  145 \\
\midrule
Harvard & Cambridge & 32 & \textbf{8530} \\
ubc.ca & Vancouver & 1 & \textbf{224} \\
stanford.edu & Baltimore & 6 & \textbf{2842} \\
wdl.org & Baltimore & 2 & \textbf{664} \\
\midrule
\textbf{Total} & & \textbf{500} & \textbf{107,227}\\
 \bottomrule
\end{tabular}

\end{table}

\subsection{Selection}

Our goal is to develop a system based on machine learning to automatically analyze a large corpus of books of hours. To train the models, a set of annotated pages is needed. Usually, a random sample of pages is selected to be annotated. However, the images of pages in the full collection show a great variety in their appearance, due to the difference in manuscripts' layout and to different conservation and digitization conditions (color scale, double-page, etc.). Therefore, in order to train a model with a high generalization capacity, we wanted to select a subset which would adequately represent this diversity.

Within a given manuscript, different types of pages share a similar layout, e.g. calendar pages, full-text pages, and illustrations. Within the corpus, some manuscripts have a similar layout and some are very different from all the others.  These rare layouts are important in order to train a system with good generalization capacity on all image types, even the rarest ones. For this reason, it was not appropriate to select a random sample among the whole corpus. We defined a strategy to select a sample reflecting the diversity of the images in the corpus. This strategy consists of three steps: automatic page classification, clustering of the pages, and selection.
\subsubsection{Automatic page classification}
This step aims at identifying pages for which no text recognition will be performed and therefore no layout analysis is needed. All pages are first automatically classified into seven classes (\textit{binding}, \textit{white page}, \textit{calendar}, \textit{miniature}, \textit{miniature-and-text}, \textit{text-with-miniature} and \textit{full-page text}) with a classifier based on deep neural networks\cite{icdar_horae2019}. We remove pages classified as \textit{binding} and \textit{white page}. After this filtering step, 92,512 images are kept.
In order to reduce the redundancy, we keep only two images of each class for each manuscript.
At this stage, 5,738 different pages are selected from the full corpus of  107,227 pages.

\subsubsection{Clustering of the pages}
The goal of this clustering step is twofold: first to group similar pages so that only one image is selected among them and second to detect pages that have a rare layout ("outliers").

In order to obtain a clustering of the 5,738 pages, we applied HDBSCAN\cite{McInnes2017}, a density-based hierarchical clustering algorithm, to all the images represented as a vector composed of the raw pixels values from a 64 × 64 sub-resolution image with the 3 color channels concatenated (12,288 values in total per image). The clustering was performed over the pages with a \textit{min\_cluster\_size} of 3 and a Euclidean metric. We tested different values for the \textit{min\_cluster\_size}. With a value equal to 4, every single image was defined as "outlier". On the contrary, with a value of 2, almost all images were clustered 2 by 2. Therefore we chose 3 is a good compromise between having too many clusters and no cluster.

The clustering algorithm HDBSCAN does at once the clustering and detection of the outliers (images of a rare type). It defined 141 different clusters and it labeled ca. 2,200 images as "outliers" (they have not been grouped with one of the 141 clusters). Both groups are used for the selection.

\subsubsection{Selection step}
We decided to select 600 images to be annotated. This number was estimated in relation to the amount of annotation effort we wanted to spent on this task (around 60 hours). We first kept the centroids of each cluster. These 141 images correspond to the most frequent layouts in the dataset. To deal with the rare layout types, we selected the 459 images with the highest "outlier score", a value computed by the clustering algorithm.

\begin{table}[ht]
\caption{Number of pages at different stages of the selection process.}
\label{tab:sampling_sizez}
\centering
\begin{tabular}{l r }
\toprule 
Processing step   & Sample size (pages)\\
\midrule 
Full HORAE corpus (500 manuscripts) & 107,227  \\
Filtering page classes & 92,512 \\
Sample based on page classes & 5,738 \\
Clustering-based selection & 600 \\
Annotated images & 557 \\
\bottomrule 
\end{tabular}

\end{table}

\begin{figure}
    \centering
    \includegraphics[width=0.55\textwidth]{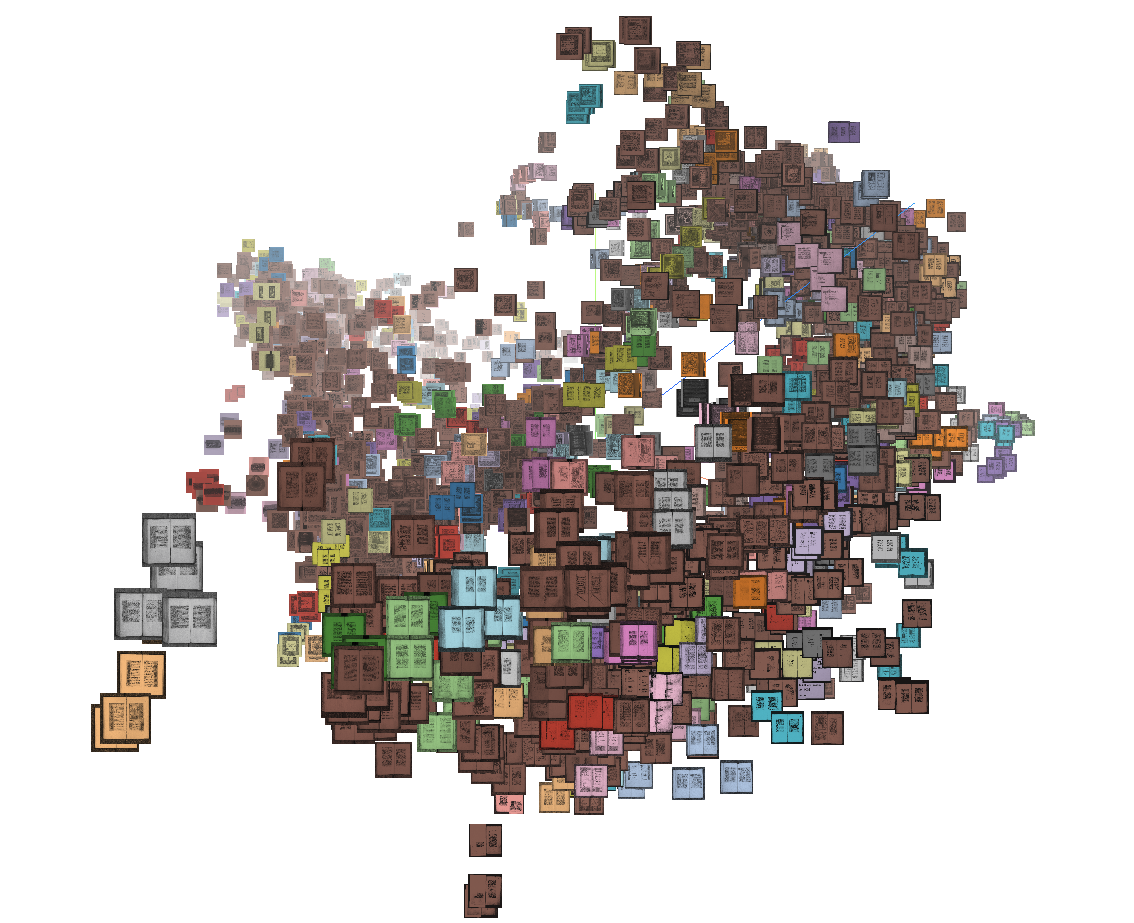}
    \caption{Visualization of the clustered pages in the 3D-space. A PCA has been performed over the 5,738 images. We show here the result of the clustering step: the pages belonging to the same cluster are highlighted in the same color. We can see that the pages that share a similar layout are assigned the same cluster.
    For example, the three images in yellow (on the left of the image) belong to the same cluster. During the selection step, only one image within this cluster of three images is kept. The images highlighted in brown are the ones defined as outliers. See \url{https://page-projector.horae.digital/}.}
    \label{fig:embedding_projector}
\end{figure}

Visualization can be performed over all the clustered images (5,738 pages). Figure \ref{fig:embedding_projector} shows the visualization of the pages in a 3D space after applying a  Principal Component Analysis (PCA) using Embedding Projector \cite{Smilkov2016}.

\subsection{Annotation process}

The 600 selected images were annotated by three annotators using Transkribus\footnote{\url{https://transkribus.eu/Transkribus/}}. They annotated the main structural elements: pages, decorated borders, text regions, and text lines, miniatures, initials, and inline decorative elements. The manually annotated structures are rectangular shapes, except for \textbf{text lines}, which were automatically detected by Transkribus' CITlab layout analysis tool within \textbf{text regions}, then corrected if necessary. The output of this annotation process are PAGE XML files in which we have the coordinates and tags for all shapes; there is one PAGE XML file per annotated image.


\begin{figure}
    \centering
    \includegraphics[width=0.50\textwidth]{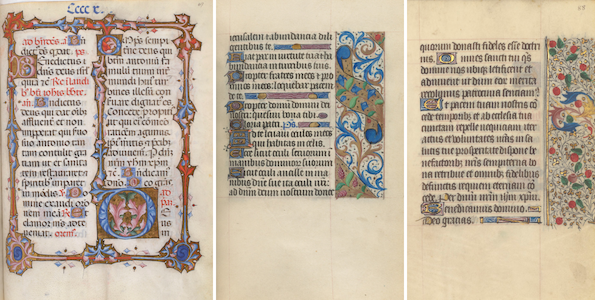}
    \caption{Examples of decorated borders.}
    \label{fig:borders}
\end{figure}

Annotators worked according to the following order and definitions:

\textbf{Page}: one for each actual page of the manuscript present in the image (one or two);

\textbf{Miniature}: all illustrations that are not part of the borders. We defined the borders as decorations that frame the text region, whether they enclose it completely or are only on some sides.

\textbf{Border elements} are divided in: \textbf{illustrated\_border} for miniatures and meaningful representations; \textbf{decorated\_border} for ornamental border decorations; \textbf{border\_text} for text regions that are located in the borders. The distinction between illustrated and decorated borders is not always an easy one; our annotation guidelines distinguished between ornamental elements which depict a scene or identifiable character and those that do not. For example, a border element depicting a saint would be annotated as an illustrated\_border, while one depicting flowers or random animals would not.

\textbf{Initials} are divided along the same principles as the border elements: \textbf{simple\_initial} for those differing from the body of the text by their ink color and/or size (some images are in black and white); \textbf{decorated\_initial} for initials that are decorated only with purely ornamental, not iconographic elements; and \textbf{historiated\_initial} for the usually bigger initials whose decoration contains an iconographic element and depicts a scene or a character, carrying a meaning beyond the alphabetical letter. The distinction is similar to the one between decorated and illustrated borders.

Other decorations inside the body of the text are annotated with: the self-explaining \textbf{line\_filler} tag; the \textbf{music\_notation} tag for musical notation; \textbf{ornamentation} tag for decorations within the text that do not fit any of the other tags.

\begin{figure}
    \centering
    \includegraphics[width=0.48\textwidth]{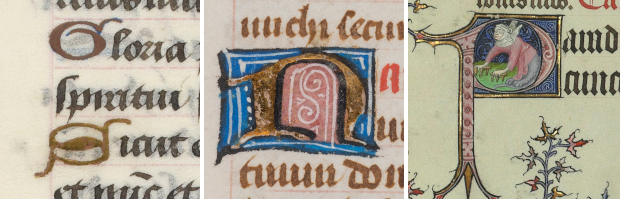}
    \caption{Three types of ornamental initials : simple initials, decorated initials and historiated initials.}
    \label{fig:initials}
\end{figure}


Because in annotating the books of hours pages there is no ground-truth we can rely on to measure the correctness of the annotation, we chose to evaluate instead the consistency of the annotations made by our three annotators. In order to do so, we added to the 200 images they each had to annotate 10 images that all three of them annotated, to evaluate the inter-annotator agreement. \\
The following table presents the results of this evaluation using the Intersection-over-Union (IoU) metric. In the absence of ground truth, we compared each annotator's boxes against the two others'.

\begin{table}[ht]

\caption{Inter-annotator agreement for the three annotators (A1, A2, A3) using the Intersection-over-Union metrics for the different annotation types.}
\label{tab:interannotator_agreement}
\centering
\begin{tabular}{l c c c c}
\toprule 
  &  A1/A2  &  A2/A3  &  A3/A1  &
   average \\
\midrule 
 page   & 0.94 & 0.85 & 0.85 & 0.88\\ 
 text region   & 0.95 & 0.95 & 0.95 & 0.95\\ 
 border\_text   & 0.95 & 0.95 & 1.0 & 0.97\\
 text line   & 0.95 & 0.88 & 0.87 & 0.90\\
 miniature   & 0.93 & 0.98 & 0.93 & 0.95\\
 decorated\_border   & 0.90 & 0.90 & 0.85 & 0.88\\
 illustrated\_border   & 0.92 & 0.97 & 0.91 & 0.93\\
 simple\_initial    & n/a & n/a & n/a & n/a\\
 decorated\_initial   & 0.92 & 0.89 & 0.90 & 0.90\\
 historiated\_initial   & 0.99 & 0.98 & 0.98 & 0.98\\
 line\_filler   & n/a & n/a & n/a & n/a\\
 music\_notation   & n/a & n/a & n/a & n/a\\
 ornamentation   & n/a & n/a & n/a & n/a\\
\midrule
 global   & 0.94 & 0.93 & 0.92 & \textbf{0.93}\\
\bottomrule 
\end{tabular}

\end{table}

The elements on which inter-annotator agreement is the lowest are the pages and the decorated borders. A visual examination of the relevant images shows that for these pages, the annotators had to choose between including some background the page zone, or excluding some of the actual pages from that zone in order to draw a rectangular shape, and the choices they made in that regard explain the IoU score. For the decorated borders, the differences mainly come from whether or not some decorations stemming from initials were identified as border decorations.\\
The IoU scores are overall very satisfactory, and we can, therefore, assume that our annotated dataset makes for good training data.

\subsection{Corpus statistics}

Among the 600 selected pages, some had been incorrectly classified, such as bookbinding, or images containing technical or bibliographical information, or signaling issues during the digitization process.

Since the selection process aimed at including infrequent images, it picked up more abnormal images that were not relevant for this structural annotation than a random selection would have. Moreover, we also excluded from this annotation process the calendar pages, opting to deal with them separately due to their tabular structure. The final corpus encompasses 557 annotated images\footnote{\url{https://github.com/oriflamms/HORAE/}}.\\

The annotations are distributed as follows: 557 images, 797 pages, 843 text regions (including 51 border\_text), 12,512 text lines, 284 miniatures, 892 decorated\_borders, 118 illustrated\_borders, 2776 decorated\_initials, 551 simple\_initials, 22 historiated\_initials, 1112 line\_fillers, 5 ornamentation, 4 music\_notations.

The number of ornamental border zones is greater than the number of pages because the annotators drew one box for each decorated margin (left, right, upper, lower), instead of drawing one rectangular shape covering all decorated margins, but containing also other elements, and from which we would have had to reconstruct the actual border by subtracting overlapping miniature or text zones.

\section{Automatic analysis}

\subsection{System description}
The variety and diversity of historical documents and the small amount of annotated data available prevent us to use any off-the-shelf network designed for analyzing more classical document images. When working with such documents, one would use a document analysis solution that is flexible and able of generalization. In this regard, we opted for the network \textit{dhSegment} \cite{oliveiraseguinkaplan2018dhsegment} suggested by \textit{Oliveira et al.} to run an automatic analysis over the data presented in the previous section. The network has shown competitive results over different tasks of historical document processing. It presents several advantages like working with a little amount of training data, a reduced training time, and it addresses various tasks such as baseline extraction and layout analysis. In addition, the implementation of the approach and different post-processing blocks are open-source.

The network is a Fully Convolutional Neural network and consists of a contracting path based on a ResNet-50 architecture and an expanding path that outputs full input resolution feature maps.

\subsection{Experiments}
In order to analyze our data, two experiments are performed. Each experiment consists of line detection and a layout analysis using \textit{dhSegment}. All the experiments are tested over the same dataset of 30 pages selected from the annotated dataset.

The two modules of the first experiment are trained over 220 images with 7 validation images. Both modules are trained for 30 epochs. The modules of the second experiment are trained for 60 epochs over 510 images with 17 validation images taken from the annotated data. For both experiments, the input images are not resized but patches of size 400 × 400 are used in order to run batch training.

After both experiments, we apply a post-processing step consisting in creating rectangles around the predictions. First, a threshold binarization is applied to the prediction for each class then, for each binary image, the contours of the connected component are detected and all the pixels of the bounding rectangle are assigned to the class.

\subsection{Results}
The performances for these semantic segmentation tasks are measured using a pixel-wise Intersection-over-Union (IoU).

\subsubsection{Line detection}
For this task, the pixels of the documents can be assigned either to the text class or to the background.
To obtain the binary masks, a step of thresholding is applied for each class to the feature maps. The threshold value has been optimized on the validation set of the first experiment (7 images) as presented in Table \ref{tab:thresholds} and the threshold value $t = 0.7$ was chosen.

As for post-processing, the connected components smaller than a given number of a pixel are removed. After several experiments, we chose the value of $min\_cc = 50$ pixels and we observed that the impact of changing the value on the results is low. Finally, the bounding rectangles for each connected component are detected.

The results obtained by the two models are presented in  Table \ref{tab:results}. Training with more data did not show a notable gain on the results. This results indicated that the neural network can still be made more complex and optimized to benefit from more data and improve the extraction results.

\subsubsection{Layout analysis}
For this second task, the pixels of the documents can be assigned either to one of the following classes: decorated border, illustrated border, miniature, text region, ornamentation, historiated initial, decorated initial, simple initial, line-filler, and background. In this experiment, the threshold chosen is $t = 0.5$ which is the best value obtained when optimizing this parameter on the validation set as presented in Table \ref{tab:thresholds}.  As for the line detection task, the connected components smaller than the default value of $min\_cc = 50$ pixels are removed and the bounding rectangles are detected.

The results for this layout analysis task are presented in Table \ref{tab:results}. The obtained IoU is lower than for the line segmentation task due to the greater number of classes involved in this task. Training with more data shows more gain than for the first task, however, this gain is still low. We can expect that more complex neural networks would improve with more data.

\begin{table}
  \caption{Optimization of the threshold parameter on the validation set for the line detection and layout analysis tasks.}
  \label{tab:thresholds}
  \begin{tabular}{cccc}
    \toprule
    Task&Threshold&IoU&\makecell{IoU with \\ post-processing}\\
    \midrule
    \multirow{4}{*}{Line detection}&
    0.3 & 0.86 & 0.84 \\
     &   0.5 & 0.87 & 0.87 \\
     &  \textbf{0.7} & \textbf{0.84} & \textbf{0.88}\\
     &   0.9 & 0.78  & 0.87 \\ \midrule
    \multirow{4}{*}{Layout Analysis} &
    0.3 & 0.39  & 0.63\\
       &\textbf{0.5} & \textbf{0.73} & \textbf{0.75}\\
       &0.7 & 0.72 & 0.74\\
       &0.9 & 0.66 & 0.68\\
  \bottomrule
\end{tabular}
\end{table}

\begin{table}
  \caption{Results of the line detection and layout analysis experiments for different training sizes.}
  \label{tab:results}
  \begin{tabular}{cccc}
    \toprule
    Task&Training size&IoU&\makecell{IoU with \\ post-processing}\\
    \midrule
    \multirow{2}{*}{Line detection} & 220 & 0.86 & 0.88\\
    & 510 & 0.87 & 0.88\\ \midrule
        \multirow{2}{*}{Layout Analysis} & 220 & 0.69 & 0.71\\
    & 510 & 0.71 & 0.72\\

  \bottomrule
\end{tabular}
\end{table}

\subsection{Visualization}
\begin{figure*}
    \centering
    \includegraphics[width=0.27\textwidth] {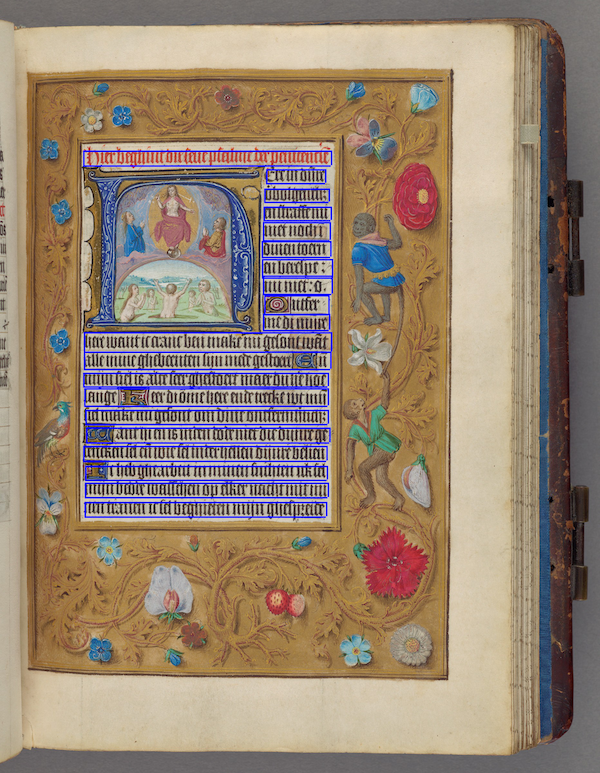}
    \includegraphics[width=0.27\textwidth]{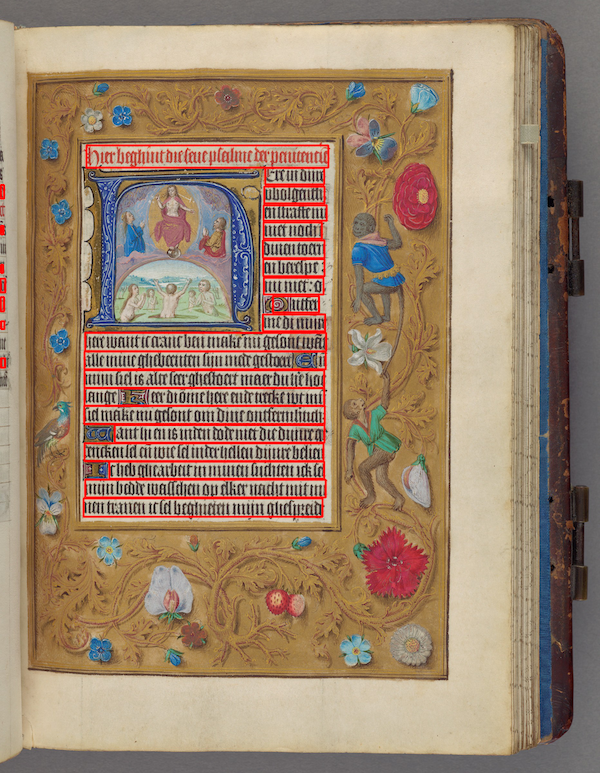}
    \includegraphics[width=0.22\textwidth]{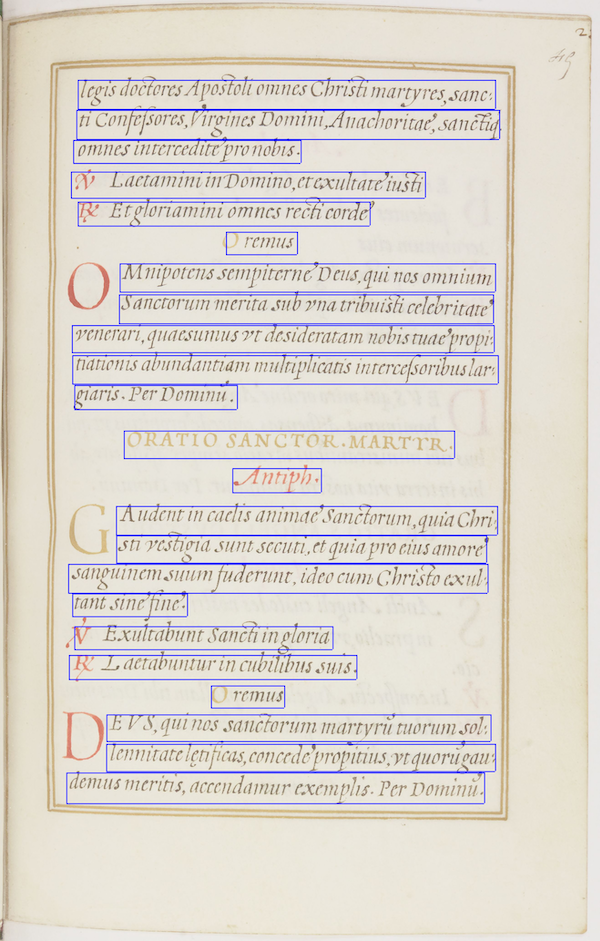}
    \includegraphics[width=0.22\textwidth]{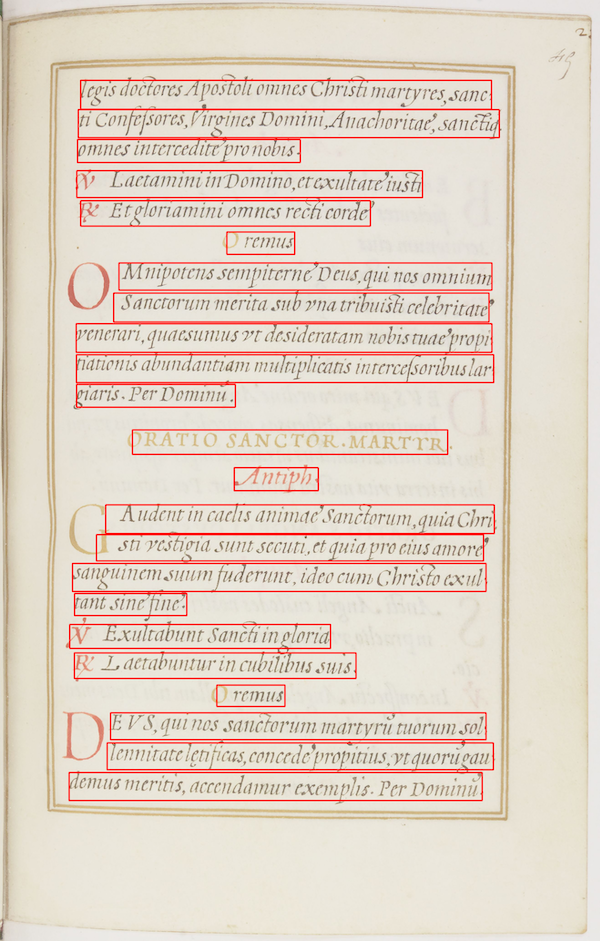}
    \caption{Examples of line segmentation on test pages. On the left, the  manuscript pages with the ground-truth and on the right the pages with the classes predicted by the model.}
    \label{fig:results_line_seg}
\end{figure*}
\begin{figure*}
    \centering
    \includegraphics[width=0.32\textwidth]{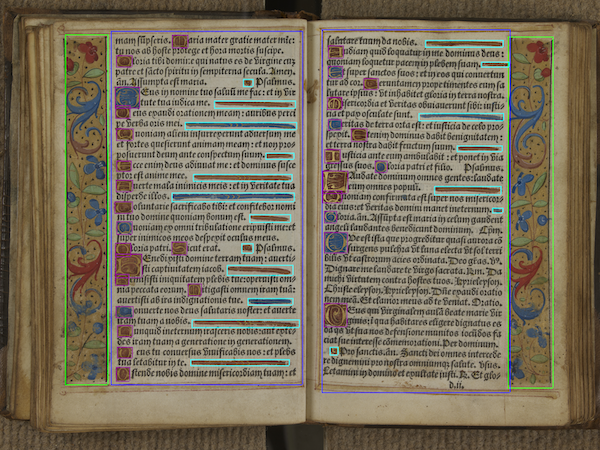}
    \includegraphics[width=0.32\textwidth]{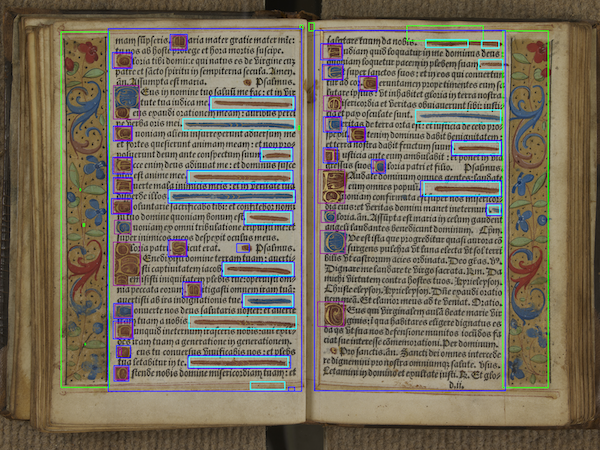}
    \includegraphics[width=0.17\textwidth]{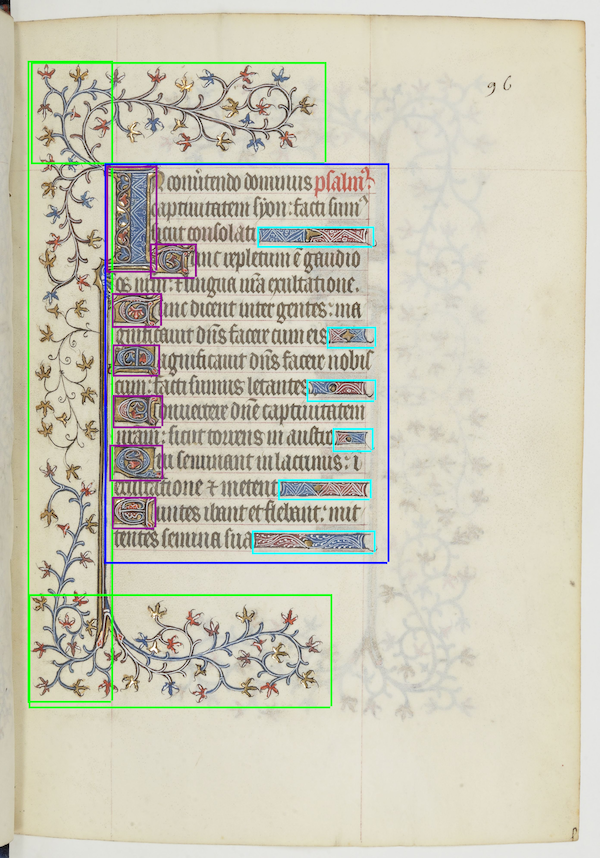}
    \includegraphics[width=0.17\textwidth]{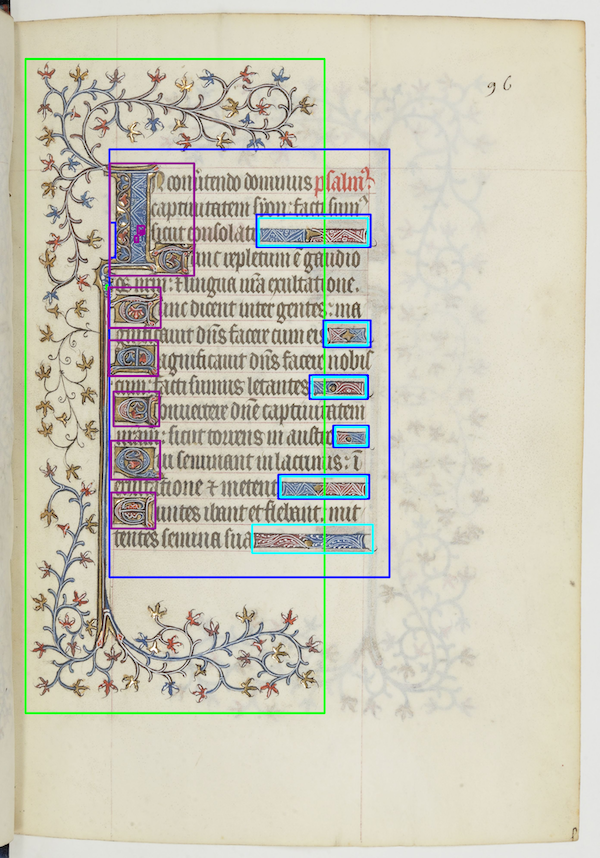}
    \caption{Examples of layout segmentation on test pages. On the left, the  manuscript pages with the ground-truth and on the right the pages with the classes predicted by the model.}
    \label{fig:results_layout_seg}
\end{figure*}

Figures \ref{fig:results_line_seg} and \ref{fig:results_layout_seg} show images that have been analyzed. On the first figure, the text lines are well detected and the boxes are similar to the ones wanted. When considering the textual content of a page, it is reasonable to include the initials in the text lines. Therefore, during the annotation process, some initials have been annotated as part of text lines as one can see on the top example. However, depending on the future application of the extracted text lines, it can be more appropriate to exclude them. This ambiguity has been partly corrected by the model since the majority of the initials are not included in the detected text lines.

The second figure shows the result of the layout analysis applied to two images. The results are also satisfying however some problems appear when creating the bounding boxes. One can see the problem on the second example where a big green rectangle has been drawn, including some text, instead of having different small green rectangles around the decorated borders.

\section{Conclusion}

In this paper, we introduce HORAE, a new dataset of annotated pages selected from a large number of book of hours. The corpus has been selected to include a large variety of type of pages using clustering and outliers detection. The layout of the pages has been fully manually annotated both for the text and for the elements of decoration. A state-of-the-art automatic segmentation system based on the deep neural network has been trained and first reference results are reported. The segmentation results are already satisfactory but the size of the dataset allows for improvement, for example with more complex neural networks.

\subsection*{Acknowledgments}
This work benefited from the support of the project HORAE ANR-17-CE38-0008 of the French National Research Agency (ANR) and from the project "Horae Pictavenses: origines et provenances des manuscrits poitevins étudiés dans le texte et l'image" funded by Equipex Biblissima.
\bibliographystyle{ACM-Reference-Format}
\bibliography{references}


\begin{thebibliography}{00}


\ifx \showCODEN    \undefined \def \showCODEN     #1{\unskip}     \fi
\ifx \showDOI      \undefined \def \showDOI       #1{{\tt DOI:}\penalty0{#1}\ }
  \fi
\ifx \showISBNx    \undefined \def \showISBNx     #1{\unskip}     \fi
\ifx \showISBNxiii \undefined \def \showISBNxiii  #1{\unskip}     \fi
\ifx \showISSN     \undefined \def \showISSN      #1{\unskip}     \fi
\ifx \showLCCN     \undefined \def \showLCCN      #1{\unskip}     \fi
\ifx \shownote     \undefined \def \shownote      #1{#1}          \fi
\ifx \showarticletitle \undefined \def \showarticletitle #1{#1}   \fi
\ifx \showURL      \undefined \def \showURL       #1{#1}          \fi
\providecommand\bibfield[2]{#2}
\providecommand\bibinfo[2]{#2}
\providecommand\natexlab[1]{#1}

\bibitem[\protect\citeauthoryear{??}{hor}{}]%
        {horae}
\bibinfo{booktitle}{{\em Heures : Reconnaissance de l'écriture manuscrite,
  catégorisation automatique, éditions – HORAE}}.
\newblock
\newblock
\shownote{\url{https://anr.fr/Projet-ANR-17-CE38-0008}, last accessed August,
  2019.}


\bibitem[\protect\citeauthoryear{Ares~Oliveira, Seguin, and
  Kaplan}{Ares~Oliveira et~al\mbox{.}}{2018}]%
        {oliveiraseguinkaplan2018dhsegment}
\bibfield{author}{\bibinfo{person}{Sofia Ares~Oliveira},
  \bibinfo{person}{Benoit Seguin}, {and} \bibinfo{person}{Frederic Kaplan}.}
  \bibinfo{year}{2018}\natexlab{}.
\newblock \showarticletitle{dhSegment: A generic deep-learning approach for
  document segmentation}. In \bibinfo{booktitle}{{\em International Conference
  on Frontiers in Handwriting Recognition}}.
\newblock


\bibitem[\protect\citeauthoryear{Boros, Toumi, Rouchet, Abadie, Stutzmann, and
  Kermorvant}{Boros et~al\mbox{.}}{2019}]%
        {icdar_horae2019}
\bibfield{author}{\bibinfo{person}{Emanuela Boros}, \bibinfo{person}{Alexis
  Toumi}, \bibinfo{person}{Erwan Rouchet}, \bibinfo{person}{Bastien Abadie},
  \bibinfo{person}{Dominique Stutzmann}, {and} \bibinfo{person}{Christopher
  Kermorvant}.} \bibinfo{year}{2019}\natexlab{}.
\newblock \showarticletitle{Automatic page classification in a large collection
  of manuscripts based on the International Image Interoperability Framework}.
  In \bibinfo{booktitle}{{\em International Conference on Document Analysis and
  Recognition}}.
\newblock


\bibitem[\protect\citeauthoryear{Fischer, Frinken, Forn{\'{e}}s, and
  Bunke}{Fischer et~al\mbox{.}}{2011}]%
        {Fischer2011}
\bibfield{author}{\bibinfo{person}{Andreas Fischer}, \bibinfo{person}{Volkmar
  Frinken}, \bibinfo{person}{Alicia Forn{\'{e}}s}, {and} \bibinfo{person}{Horst
  Bunke}.} \bibinfo{year}{2011}\natexlab{}.
\newblock \showarticletitle{{Transcription alignment of Latin manuscripts using
  hidden Markov models}}. In \bibinfo{booktitle}{{\em Workshop on Historical
  Document Imaging and Processing}}.
\newblock


\bibitem[\protect\citeauthoryear{Fischer, Keller, Frinken, and Bunke}{Fischer
  et~al\mbox{.}}{2012}]%
        {Fischer2012}
\bibfield{author}{\bibinfo{person}{Andreas Fischer}, \bibinfo{person}{Andreas
  Keller}, \bibinfo{person}{Volkmar Frinken}, {and} \bibinfo{person}{Horst
  Bunke}.} \bibinfo{year}{2012}\natexlab{}.
\newblock \showarticletitle{{Lexicon-free handwritten word spotting using
  character HMMs}}.
\newblock \bibinfo{journal}{{\em Pattern Recognition Letters\/}}
  \bibinfo{volume}{{33}, 7} (\bibinfo{year}{2012}), \bibinfo{pages}{934--942}.
\newblock


\bibitem[\protect\citeauthoryear{Grana, Borghesani, and Cucchiara}{Grana
  et~al\mbox{.}}{2009}]%
        {Grana2009}
\bibfield{author}{\bibinfo{person}{Costantino Grana}, \bibinfo{person}{Daniele
  Borghesani}, {and} \bibinfo{person}{Rita Cucchiara}.}
  \bibinfo{year}{2009}\natexlab{}.
\newblock \showarticletitle{{Picture extraction from digitized historical
  manuscripts}}. In \bibinfo{booktitle}{{\em International Conference on Image
  and Video Retrieval}}.
\newblock


\bibitem[\protect\citeauthoryear{McInnes, Healy, and Astels}{McInnes
  et~al\mbox{.}}{2017}]%
        {McInnes2017}
\bibfield{author}{\bibinfo{person}{Leland McInnes}, \bibinfo{person}{John
  Healy}, {and} \bibinfo{person}{Steve Astels}.}
  \bibinfo{year}{2017}\natexlab{}.
\newblock \showarticletitle{{HDBSCAN}: Hierarchical density based clustering}.
\newblock \bibinfo{journal}{{\em The Journal of Open Source Software\/}}
  \bibinfo{volume}{{2}, 11} (\bibinfo{date}{mar} \bibinfo{year}{2017}).
\newblock
\showDOI{%
\url{http://dx.doi.org/10.21105/joss.00205}}


\bibitem[\protect\citeauthoryear{Mehri, H{\'e}roux, Mullot, Moreux,
  Co\"{u}asnon, and Barrett}{Mehri et~al\mbox{.}}{2017}]%
        {Mehri_2017}
\bibfield{author}{\bibinfo{person}{Maroua Mehri}, \bibinfo{person}{Pierre
  H{\'e}roux}, \bibinfo{person}{R{\'e}my Mullot},
  \bibinfo{person}{Jean-Philippe Moreux}, \bibinfo{person}{Bertrand
  Co\"{u}asnon}, {and} \bibinfo{person}{Bill Barrett}.}
  \bibinfo{year}{2017}\natexlab{}.
\newblock \showarticletitle{{HBA 1.0}: A Pixel-based Annotated Dataset for
  Historical Book Analysis}. In \bibinfo{booktitle}{{\em Workshop on Historical
  Document Imaging and Processing}}.
\newblock


\bibitem[\protect\citeauthoryear{Smilkov, Thorat, Nicholson, Reif,
  Vi{\'{e}}gas, and Wattenberg}{Smilkov et~al\mbox{.}}{2016}]%
        {Smilkov2016}
\bibfield{author}{\bibinfo{person}{Daniel Smilkov}, \bibinfo{person}{Nikhil
  Thorat}, \bibinfo{person}{Charles Nicholson}, \bibinfo{person}{Emily Reif},
  \bibinfo{person}{Fernanda~B. Vi{\'{e}}gas}, {and} \bibinfo{person}{Martin
  Wattenberg}.} \bibinfo{year}{2016}\natexlab{}.
\newblock \showarticletitle{{Embedding Projector: Interactive Visualization and
  Interpretation of Embeddings}}. In \bibinfo{booktitle}{{\em NIPS 2016
  Workshop on Interpretable Machine Learning in Complex Systems}}.
\newblock


\bibitem[\protect\citeauthoryear{Stutzmann}{Stutzmann}{2019}]%
        {Stutzmann19}
\bibfield{author}{\bibinfo{person}{Dominique Stutzmann}.}
  \bibinfo{year}{2019}\natexlab{}.
\newblock \showarticletitle{R\'esistance au changement ? Les \'ecritures des
  livres d'heures dans l'espace fran\c{c}ais (1200-1600)}. In
  \bibinfo{booktitle}{{\em 'Change' in Medieval and Renaissance Scripts and
  Manuscripts. Proceedings of the 19th Colloquium of the Comité international
  de paléographie latine (Berlin, 16-18 September, 2015)}}. Brepols, Turnhout,
  \bibinfo{pages}{101--120}.
\newblock


\bibitem[\protect\citeauthoryear{Stutzmann, Currie, Daille, Hazem, and
  Kermorvant}{Stutzmann et~al\mbox{.}}{2019}]%
        {dh_stutzmann2019}
\bibfield{author}{\bibinfo{person}{Dominique Stutzmann}, \bibinfo{person}{Jacob
  Currie}, \bibinfo{person}{Béatrice Daille}, \bibinfo{person}{Amir Hazem},
  {and} \bibinfo{person}{Christopher Kermorvant}.}
  \bibinfo{year}{2019}\natexlab{}.
\newblock \showarticletitle{Integrated {DH.} Rationale of the {HORAE} Research
  Project}. In \bibinfo{booktitle}{{\em Digital Humanities}} (2019-07-09).
\newblock


\bibitem[\protect\citeauthoryear{Yang, Pintus, Gobbetti, and Rushmeier}{Yang
  et~al\mbox{.}}{2017}]%
        {Yang2017}
\bibfield{author}{\bibinfo{person}{Ying Yang}, \bibinfo{person}{Ruggero
  Pintus}, \bibinfo{person}{Enrico Gobbetti}, {and} \bibinfo{person}{Holly
  Rushmeier}.} \bibinfo{year}{2017}\natexlab{}.
\newblock \showarticletitle{{Automatic Single Page-Based Algorithms for
  Medieval Manuscript Analysis}}.
\newblock \bibinfo{journal}{{\em Journal on Computing and Cultural Heritage\/}}
  \bibinfo{volume}{{10}, 2} (\bibinfo{year}{2017}), \bibinfo{pages}{1--22}.
\newblock


\end{thebibliography}

\end{document}